\documentclass{article}
\newcommand{\thanksnomark}[1]{%
  \begingroup
  \renewcommand{\thefootnote}{}%
  \footnotetext{*#1}%
  \endgroup
}

\usepackage{arxiv}

\usepackage[utf8]{inputenc} 
\usepackage[T1]{fontenc}    
\usepackage{hyperref}       
\usepackage{url}            
\usepackage{booktabs}       
\usepackage{amsfonts}       
\usepackage{amsmath}
\usepackage{amssymb}
\usepackage{nicefrac}       
\usepackage{microtype}      
\usepackage{lipsum}
\usepackage{graphicx}
\usepackage{algorithm}
\usepackage{algorithmic}
\usepackage{multirow}
\usepackage{xcolor}
\usepackage{tabularx}
\usepackage{caption}
\usepackage{subcaption}
\usepackage{authblk}
\usepackage{dsfont}
\graphicspath{ {./images/} }

\title{Entropy-Tree: Tree-Based Decoding with Entropy-Guided Exploration}

\author[1,3]{\textbf{Longxuan Wei}}
\author[2]{\textbf{Yubo Zhang}}
\author[1]{\textbf{Zijiao Zhang}}
\author[2]{\textbf{Zhihu Wang}}
\author[4]{\textbf{Shiwan Zhao}}
\author[1]{\\\textbf{Tianyu Huang}}
\author[5]{\textbf{Huiting Zhao}}
\author[2]{\textbf{Chenfei Liu}}
\author[1,3]{\textbf{Shenao Zhang}}
\author[1,3,*]{\textbf{Junchi Yan}}

\affil[1]{Shanghai Jiaotong University}
\affil[2]{Huawei Technologies Ltd.}
\affil[3]{Shanghai Innovation Institute}
\affil[4]{Nankai University}
\affil[5]{Nanyang Technological University}

\begin{document}

\date{}
\maketitle
\thanksnomark{Corresponding author}
\begin{center}
\texttt{\{weilongxuan189034,zijiao.zhang,tianyuhuang,olliezhang7,yanjunchi\}@sjtu.edu.cn} \\
\texttt{\{zhangyubo20,wangzhihu3,liuchenfei2\}@huawei.com} \\
\texttt{zhaosw@gmail.com \quad huiting001@e.ntu.edu.sg} \\
\vspace{1em}
\end{center}

\begin{abstract}
Large language models achieve strong reasoning performance, yet existing decoding strategies either explore blindly (random sampling) or redundantly (independent multi-sampling). We propose Entropy-Tree, a tree-based decoding method that exploits entropy as a signal for branching decisions—expanding the search tree only at positions where the model exhibits genuine uncertainty. Entropy-Tree shows superior accuracy and calibration in reasoning tasks: it achieves better pass@k than Multi-chain across multiple models and datasets, and its predictive entropy demonstrates better AUROC compared to several traditional metrics. Entropy-Tree unifies efficient structured exploration and reliable uncertainty estimation within a single decoding procedure.
\end{abstract}


\section{Introduction}

Large language models (LLMs) have demonstrated remarkable capabilities in complex reasoning tasks, from mathematical problem-solving to scientific analysis \cite{openai2024gpt4technicalreport,touvron2023llamaopenefficientfoundation,yang2024qwen2technicalreport}. The quality of generated outputs critically depends on the decoding strategy---how the model selects tokens from its predicted probability distributions. However, existing decoding methods exhibit fundamental limitations in exploring the model's decision space.

Traditional decoding methods have limitations. Deterministic methods such as greedy decoding \cite{jm3} always select the highest-probability token, leading to myopic decisions that often produce repetitive or suboptimal outputs. Beam search \cite{tillmann-ney-2003-word} maintains multiple candidate sequences but remains a pruned greedy strategy with limited diversity, biased toward ``safe'' high-probability responses. Stochastic methods including top-k \cite{radford2019language} and top-p (nucleus) sampling \cite{holtzman2020curiouscaseneuraltext} introduce randomness to enhance diversity, yet their exploration is blind---uniformly applied across all positions regardless of the model's actual uncertainty at each step. 

When faced with more challenging reasoning tasks, multiple samples may be needed to find the correct answer. Self-Consistency \cite{wang2023selfconsistencyimproveschainthought} addresses reasoning tasks by generating multiple independent samples and selecting the most frequent answer through majority voting. While effective, this approach suffers from computational redundancy: independent samples do not share prefix computations, and the exploration is undirected, potentially wasting resources on semantically redundant paths while missing critical decision points.

Moreover, traditional LLM uncertainty quantification methods are also based on multiple random sampling, using information such as output log probabilities or semantic similarity for uncertainty assessment. The randomness of sampling imposes limitations on the quantification of uncertainty. 

Recent work by Wang et al. \cite{wang20258020rulehighentropyminority} reveals a striking pattern in chain-of-thought reasoning: token entropy distributions are highly non-uniform. Approximately 80\% of tokens exhibit low entropy---the model generates them with high confidence, primarily for syntactic coherence or continuing the current reasoning thread. The remaining 20\% of high-entropy tokens serve as ``decision forks'' that steer the reasoning trajectory toward different directions. These high-entropy positions often correspond to semantically significant tokens such as logical connectives (``thus'', ``however'') and question words (``how'', ``what''). This observation suggests a principled approach: rather than exploring uniformly or randomly, we should concentrate computational resources on these high-entropy decision points where the model genuinely hesitates among alternatives.

\begin{figure}[htbp]
    \centering
    \includegraphics[width=0.7\textwidth]{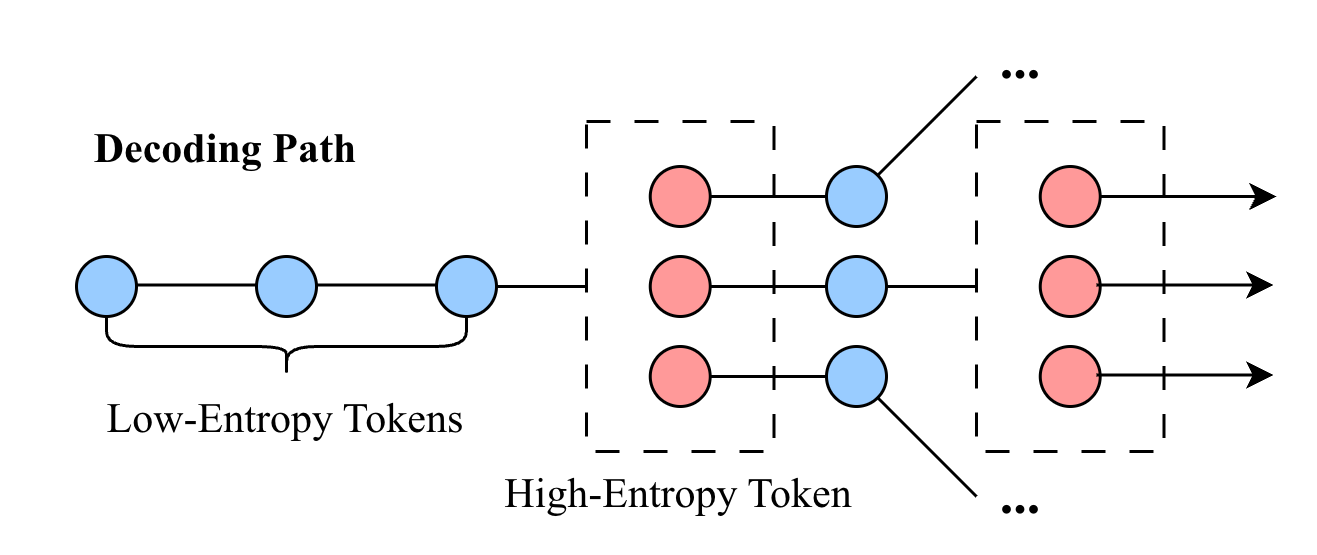}
    \caption{Entropy-Tree: Branching at high entropy tokens to form multiple decoding paths.}
    \label{fig:simple}
\end{figure}

We propose \textbf{Entropy-Tree}, a decoding strategy that utilizes high-entropy token information for tree-structured decoding. Entropy-Tree dynamically monitors the entropy of each token during the decoding process to assess the model's confidence, and branches at points where the model hesitates and where there is semantic importance, allowing the model to efficiently explore different reasoning paths. As shown in Figure~\ref{fig:simple}, whenever a high entropy token is encountered, which indicates a point of hesitation in the model's inference, Entropy-Tree will actively branch out, creating multiple inference paths. This allows the model to explore various possible subsequent outcomes.. Specifically, our contributions are as follows:
\begin{itemize}
    \item \textbf{Novel Decoding Framework}: We propose Entropy-Tree, which constructs tree-structured outputs by branching at high-entropy positions with prefix sharing, enabling structured exploration of the model's decision space.

    \item \textbf{Improved reasoning Accuracy}: Entropy-Tree demonstrated superior pass@k performance compared to random multiple sampling across the models and datasets we tested, which proves the contribution of our method to enhancing the accuracy of models in reasoning tasks.

    \item \textbf{Tree-Derived Uncertainty Quantification}: The predictive entropy calculated from the leaf nodes generated by Entropy-Tree exhibits better calibration, showing a significant improvement in AUROC compared to other metrics.
\end{itemize}

\section{Preliminary}

Let $\mathcal{V}$ denote a vocabulary of size $V$. An autoregressive language model parameterized by $\theta$ defines a probability distribution over sequences:
\begin{equation}
    P_\theta(\mathbf{y} | \mathbf{x}) = \prod_{t=1}^{T} P_\theta(y_t | \mathbf{x}, \mathbf{y}_{<t})
\end{equation}
where $\mathbf{x} = (x_1, \ldots, x_n)$ is the input prompt and $\mathbf{y} = (y_1, \ldots, y_T)$ is the generated sequence. At each timestep $t$, the model outputs a distribution $P_\theta(y_t | \mathbf{x},\mathbf{y}_{<t})$ over the vocabulary conditioned on the prompt and previously generated tokens.

The decoding strategy refers to the method of selecting tokens from the probability distribution output by the model. \textbf{Greedy decoding} \cite{jm3} directly selects the token with the highest probability:
\begin{equation}
    y_{greedy} = \arg\max_{i\in V}P_\theta(i)
\end{equation}

This method, although simple to implement, can lead to a lack of diversity in the model's generated results and is prone to getting stuck in local optima. To introduce randomness and diversity in the model generation process, researchers have proposed top-k and top-p sampling, 

\textbf{Top-k sampling} \cite{radford2019language} is shown in equation~\ref{equa:top_k}. $\mathcal{K}$ denotes the collection of the top-k tokens with the highest probability.
\begin{equation}\label{equa:top_k}
    P_{top-k}(i)=
\begin{cases}
\frac{P(i)}{\sum_{j\in\mathcal{K}}P(j)}, i\in\mathcal{K} \\
0, otherwise
\end{cases}
\end{equation}
\textbf{Top-p sampling} \cite{holtzman2020curiouscaseneuraltext} sort the tokens in descending order according to P(i) to obtain the sequence $[i_1, i_2, ..., i_\mathcal{V}]$. Then, starting from the first element of the sequence, gradually accumulate the probabilities until $\sum_{j=1}^{m}P(i_j)\ge p$. At this point, denote the set of these m tokens as $\mathcal{P}$.
\begin{equation}\label{equa:top_p}
    P_{top-p}(i)=
\begin{cases}
\frac{P(i)}{\sum_{j\in\mathcal{P}}P(j)}, i\in\mathcal{P} \\
0, otherwise
\end{cases}
\end{equation}

Top-k and top-p sampling do not simply select the token with the highest probability as the result; instead, they perform random sampling from the updated probability distribution. These methods introduce a certain level of diversity into the decoding process, but this diversity is indiscriminate, meaning that the model's performance on a single instance of a problem remains limited.

To improve the accuracy of model inference and reduce the randomness of a single sampling, \textbf{Self-Consistency} \cite{wang2023selfconsistencyimproveschainthought} samples $N$ independent reasoning paths $\{\mathbf{y}^{(1)}, \ldots, \mathbf{y}^{(N)}\}$ from the model, extracts answers $\{a^{(1)}, \ldots, a^{(N)}\}$, and selects the most frequent answer:
\begin{equation}
    \hat{a} = \arg\max_{a} \sum_{j=1}^{N} \mathds{1}[a^{(j)} = a]
\end{equation}

\section{Entropy-Tree Decoding}

\begin{figure}[htbp]
    \centering
    \includegraphics[width=1.0\textwidth]{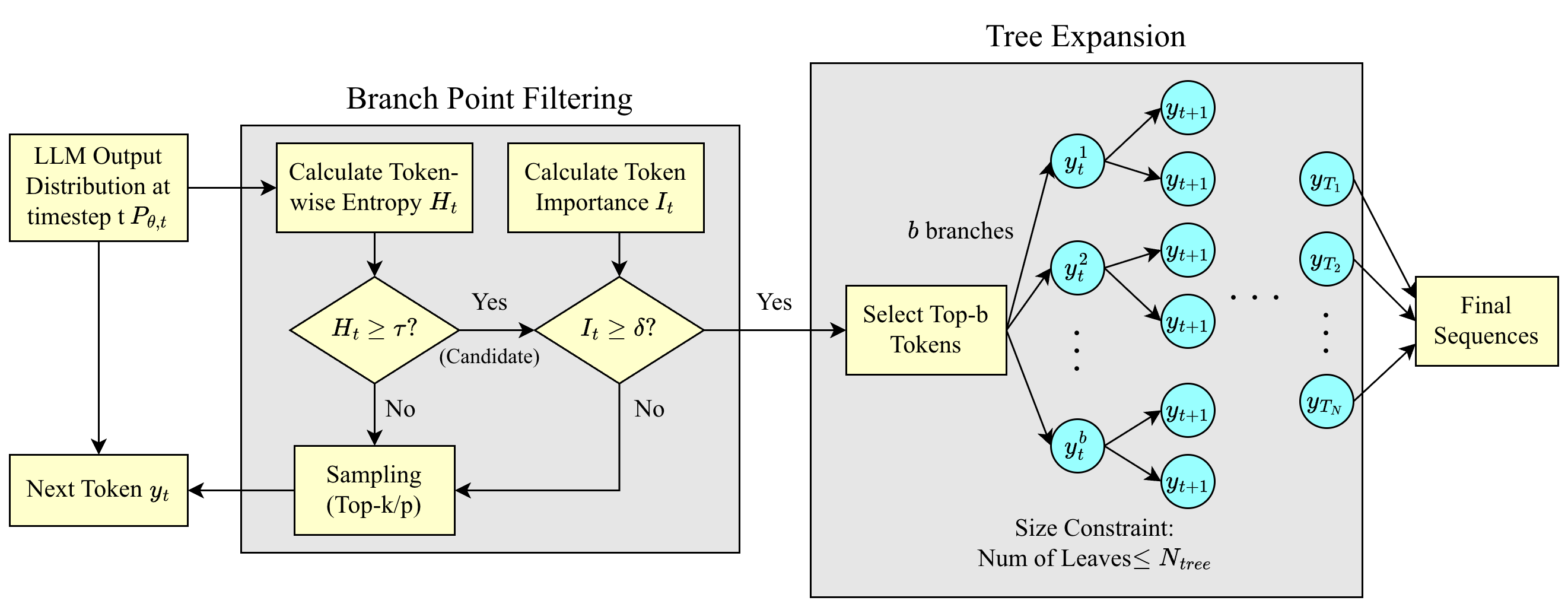}
    \caption{The complete decoding process of Entropy-Tree.}
    \label{fig:main}
\end{figure}

In this section, we will provide a detailed explanation of the Entropy-Tree decoding strategy, including the branch point filtering introduced in Section~\ref{sec:branch} and the expansion and size control described in Section~\ref{sec:tree_exp}. We will also explain in Section~\ref{sec:uncertainty} how to use the Entropy-Tree for uncertainty quantification. The complete process of the Entropy-Tree is shown in Figure~\ref{fig:main}.

\subsection{Branch Point Filtering}
\label{sec:branch}
During the LLM decoding process, we calculate the token-wise entropy based on the token probability distribution $P_{\theta,t}(i)$ as follows:
\begin{equation}
    H_t=-\sum_{i\in\mathcal{V}}p_{\theta,t}(i)\log{p_{\theta,t}(i)}
\end{equation}

Based on 80/20 rule \cite{wang20258020rulehighentropyminority}, we hypothesize that branching at high-entropy tokens can generate more exploratory and semantically diverse reasoning paths. We set an entropy threshold $\tau$. For token $y_t$ at timestep $t$, if $H_t<\tau$, a standard random sampling strategy is used (top-k and top-p sampling). If $H_t\ge\tau$, we designate this token as a candidate branching token.

Although previous research suggests that high-entropy tokens may play a decisive role in the model's reasoning process, there are also tokens with relatively low semantic importance (such as "the" and "it"). The model exhibits high entropy on these tokens due to indecision about syntactic structure, rather than encountering a critical point in the semantic or reasoning path \cite{xiong2024llmsexpressuncertaintyempirical}. Thus, we further filter the candidate branching tokens based on self-attention.

DRAGIN \cite{su-etal-2024-dragin} uses the maximum attention value to represent the importance of a token. We adopt a similar approach, and the attention matrix is calculated as follows:
\begin{equation}
    A=softmax(mask(\frac{QK^T}{\sqrt{d_k}}))
\end{equation}
$Q$ is the query matrix, $K$ is the key matrix, $d_k$ denotes the dimensionality of a key vector. The importance score for token $y_t$ is identified by the highest $A_{t,j}$ for all $j<t$:
\begin{equation}
    I_t=\max_{j=1,\ldots,t-1}A_{t,j}
\end{equation}

For the candidate branching tokens previously selected based on token-wise entropy, since the specific token to choose has not yet been determined, we calculate the importance value $I_t$ of the token $y_t$ with the highest probability. Here, we also set an importance threshold $\delta$. If $I_t<\delta$, no branching is performed; if $I_t\ge\delta$, it is determined to branch at the current token.

\subsection{Tree Expansion and Size Control}
\label{sec:tree_exp}

Once a token is identified as a branching token, random sampling is no longer used to generate a single token. Instead, the top $b$ tokens with the highest probabilities are selected as new branches. Sub-branches continue to be generated following the same process, and whenever a branch token is detected, branching will continue. We use a breadth-first search algorithm to expand the tree, ensuring that nodes with lower depth are prioritized for expansion. This ensures that all leaf nodes are completed after a certain number of branching steps.

As the number of branch nodes increases, the size of the tree grows exponentially. To prevent scale explosion, we limit the total number of leaf nodes in the tree to $N_{tree}$. Once the number of leaf nodes equals $N_{tree}$, branching will cease. The remaining branches will revert to sampling mode until an EOS token is encountered.

\subsection{Uncertainty Quantification in Entropy-Tree}
\label{sec:uncertainty}

For inference tasks, we extract the answers from all the leaf nodes and calculate the probability distribution $p(a | x)$ of the answers. We then compute the predictive entropy using the following formula:
\begin{equation}
    H=-\sum_{i=1}^{N}p(a_i|x)\log p(a_i|x)
\end{equation}

This approach is no different from the traditional method of calculating predictive entropy, except that we no longer use the answer distribution obtained from multiple random samples; instead, we use the distribution from the Entropy-Tree.

\section{Experiments}
We found that Entropy-Tree decoding not only helps improve the accuracy of answers but also enhances calibration. Therefore, we designed experiments in these two aspects.

\subsection{Experimental Setup}

\paragraph{Evaluation Metric on Accuracy.}Pass@k is a widely used evaluation metric in machine learning, particularly for assessing the performance of generative models in tasks like code generation \cite{chen2021evaluatinglargelanguagemodels}. It measures the probability that a model produces at least one correct solution out of $k$ independently generated samples for a given problem. This metric is especially valuable for quantifying a model's exploratory capability, as it rewards diversity in outputs and accounts for scenarios where multiple attempts might yield a valid result, making it ideal for evaluating creativity and robustness in AI systems.

\paragraph{Evaluation Metric on Calibration.}
The area under the receiver operator characteristic curve (AUROC) corresponds to the probability that incorrect answers have higher uncertainty scores than correct answers, and it can be used to assess the calibration of uncertainty metrics \cite{band2022benchmarkingbayesiandeeplearning}. A higher AUROC score is better, with a perfect uncertainty measurement scoring 1, while a completely random uncertainty measurement scores 0.5.

\paragraph{Baselines for Accuracy.}We compared Entropy-Tree decoding with Multi-chain, which involves the accuracy of $N$ random independent samples, to demonstrate that our method is superior to simple random sampling.

\paragraph{Baselines for Calibration.}
Predictive entropy \cite{kadavath2022languagemodelsmostlyknow} refers to the uncertainty inherent in the output distribution of a model's multiple responses. In our work, we compare the predictive entropy of the Entropy-Tree with Multi-chain. Semantic entropy \cite{kuhn2023semanticuncertaintylinguisticinvariances} measures uncertainty by performing semantic clustering on the output sequences of the model, but it may become ineffective when dealing with long text responses. Length-normalized predictive entropy \cite{malinin2021uncertaintyestimationautoregressivestructured} involves dividing the joint log-probability of each sequence by the sequence length. p(True) \cite{kadavath2022languagemodelsmostlyknow} is used to measure the model's uncertainty by asking the model again whether its previous answer was correct. Lexical Similarity \cite{fomicheva-etal-2020-unsupervised} uses the average similarity of sequences to assess uncertainty.

\paragraph{Datasets.}Entropy-Tree is suitable for reasoning datasets of a certain level of difficulty. The datasets we have selected are as follows:
\begin{itemize}
    \item \textbf{SVAMP} \cite{patel-etal-2021-nlp}: 300 elementary-level math word problems.
    \item \textbf{MATH-500} \cite{NEURIPS_DATASETS_AND_BENCHMARKS2021_be83ab3e}: 500 representative high school math problems with varying difficulty.
    \item \textbf{SciBench} \cite{wang2024scibenchevaluatingcollegelevelscientific}: 692 college-level science problems.
    \item \textbf{GPQA-main/diamond} \cite{rein2023gpqagraduatelevelgoogleproofqa}: 448/198 graduate-level physics, chemistry, biology questions.
    \item \textbf{AIME24/25}: 30 competition-level math problems each set.
\end{itemize}

\paragraph{Models.}
We used models of different scales from the Qwen2.5 series \cite{qwen2.5}, including Qwen2.5-7B-Instruct, Qwen2.5-14B-Instruct, and Qwen2.5-32B-Instruct.


\paragraph{Budget \& Fairness.}
For fair comparison, we control the number of leaves/samples: Entropy-Tree with $N=20$ leaves vs. Self-Consistency with $N=20$ independent samples. We set the threshold for determining high entropy and high importance at the 80th percentile.

\subsection{Case Study}

\begin{figure}[htbp]
    \centering
    \includegraphics[width=1.0\textwidth]{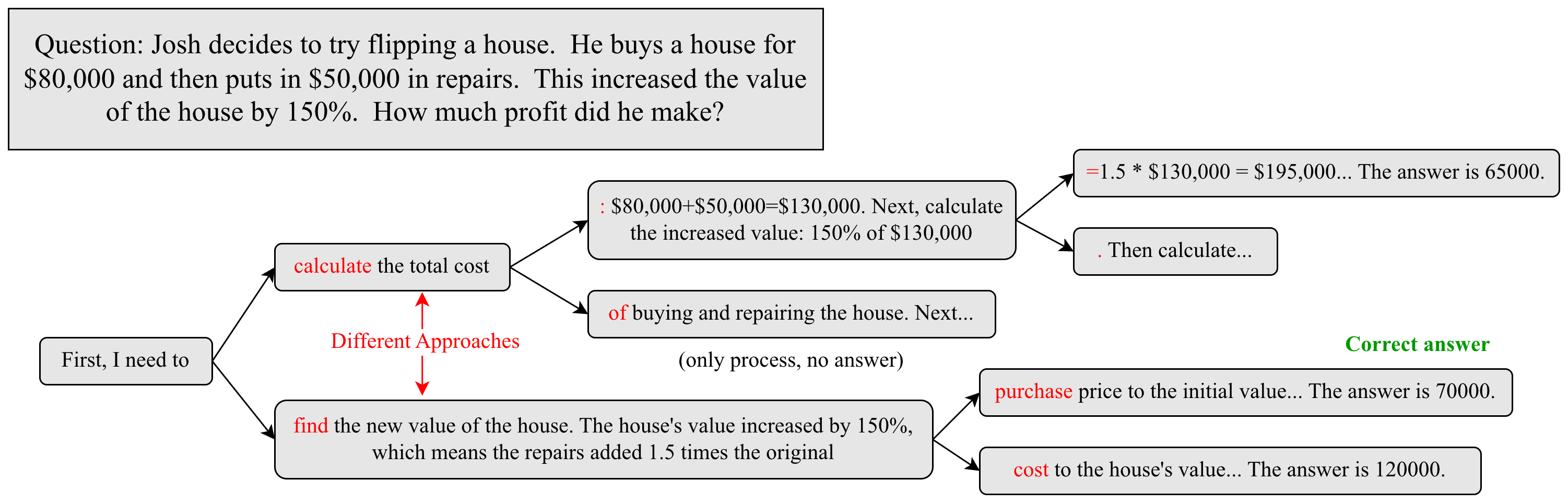}
    \caption{Decoding example of Entropy-Tree.}
    \label{fig:case}
\end{figure}

Entropy-Tree branches at key tokens, allowing for different approaches and problem-solving processes for the same issue. As shown in Figure~\ref{fig:case}, LLM identified two different methods at the first branching. In the subsequent branches corresponding to each method, different scenarios emerged, such as some leaf nodes lacking a complete calculation process and some leaf nodes having incorrect calculation results. However, because certain nodes retained the correct calculation prefix, leaf nodes that ultimately arrived at the correct answer appeared.

\subsection{Main Results}

\begin{figure}[htbp]
    \centering
    \includegraphics[width=0.6\textwidth]{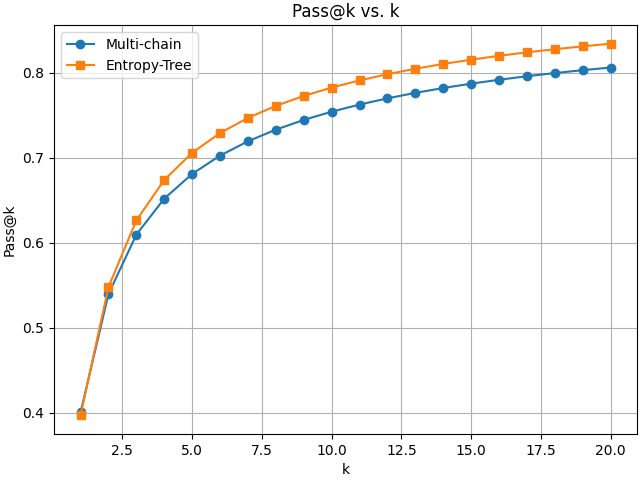}
    \caption{The pass@k curve of Qwen2.5-7B-Instruct on MATH-500.}
    \label{fig:passk_exam}
\end{figure}

The curve shown in Figure~\ref{fig:passk_exam} represents the pass@k performance of Entropy-Tree under Qwen2.5-7B-Instruct model and MATH-500 dataset. Starting from $k=3$, Entropy-Tree demonstrates a significant improvement compared to Multi-chain, with the enhancement becoming more pronounced as $k$ increases. Entropy-Tree requires approximately 2/3 of the sample size ($k=13$) to achieve a pass@k value comparable to that of Multi-chain ($k=20$).

The pass@10 and pass@20 results for all models and datasets are shown in Table~\ref{tab:pass10} and Table~\ref{tab:pass20}. In most cases, Entropy-Tree demonstrates improved accuracy compared to Multi-chain. This indicates that Entropy-Tree is suitable for reasoning tasks across different domains and varying levels of difficulty. By branching into different inference paths at high-entropy points, where the model hesitates in its decision-making, the model can significantly increase its level of exploration when faced with challenging problems. This naturally leads to an improvement in pass@k.

\begin{table}[htbp]
\centering
\caption{Pass@10 comparison across models and datasets.}
\begin{tabularx}{\textwidth}{l c *{7}{X}}
\toprule
Model & Method & SVAMP & MATH-500 & SciBench & GPQA-main & GPQA-diamond & AIME24 & AIME25 \\
\midrule
\multirow{2}{*}{Qwen2.5-7B-Instruct}
& Multi-chain & 94.37\% & 75.41\% & 57.52\% & 71.56\% & 70.83\% & 9.98\% & 17.99\% \\
& \textbf{Entropy-Tree} & \textbf{94.77\%} & \textbf{78.24\%} & \textbf{58.27\%} & \textbf{72.07\%} & \textbf{74.81\%} & \textbf{11.64\%} & \textbf{23.84\%} \\
\midrule
\multirow{2}{*}{Qwen2.5-14B-Instruct}
& Multi-chain & 96.40\% & 84.78\% & 71.05\% & 72.11\% & 73.25\% & 11.66\% & 24.64\% \\
& \textbf{Entropy-Tree} & \textbf{96.62\%} & 82.96\% & \textbf{71.31\%} & \textbf{73.53\%} & \textbf{74.10\%} & \textbf{12.54\%} & \textbf{26.20\%} \\
\midrule
\multirow{2}{*}{Qwen2.5-32B-Instruct}
& Multi-chain & 95.37\% & 77.56\% & 70.58\% & 76.70\% & 78.29\% & 14.86\% & 21.34\% \\
& \textbf{Entropy-Tree} & \textbf{95.46\%} & \textbf{79.55\%} & \textbf{72.23\%} & 75.93\% & 77.98\% & \textbf{18.33\%} & \textbf{21.78\%} \\
\bottomrule
\end{tabularx}
\label{tab:pass10}
\end{table}

\begin{table}[htbp]
\centering
\caption{Pass@20 comparison across models and datasets.}
\begin{tabularx}{\textwidth}{l c *{7}{X}}
\toprule
Model & Method & SVAMP & MATH-500 & SciBench & GPQA-main & GPQA-diamond & AIME24 & AIME25 \\
\midrule
\multirow{2}{*}{Qwen2.5-7B-Instruct}
& Multi-chain & 95.67\% & 80.60\% & 65.75\% & 79.46\% & 78.79\% & 10.00\% & 23.33\% \\
& \textbf{Entropy-Tree} & \textbf{96.00\%} & \textbf{83.40\%} & \textbf{66.62\%} & \textbf{79.91\%} & \textbf{83.84\%} & \textbf{13.33\%} & \textbf{36.67\%} \\
\midrule
\multirow{2}{*}{Qwen2.5-14B-Instruct}
& Multi-chain & 97.00\% & 88.40\% & 76.88\% & 79.24\% & 78.28\% & 13.33\% & 30.00\% \\
& \textbf{Entropy-Tree} & \textbf{97.33\%} & 85.60\% & 76.88\% & \textbf{80.36\%} & \textbf{79.80\%} & 13.33\% & \textbf{33.33\%} \\
\midrule
\multirow{2}{*}{Qwen2.5-32B-Instruct}
& Multi-chain & 96.00\% & 82.40\% & 75.14\% & 83.26\% & 82.83\% & 16.67\% & 26.67\% \\
& \textbf{Entropy-Tree} & 96.00\% & \textbf{84.60\%} & \textbf{77.46\%} & 81.92\% & \textbf{83.84\%} & \textbf{23.33\%} & 26.67\% \\
\bottomrule
\end{tabularx}
\label{tab:pass20}
\end{table}

\begin{figure}[htbp]
    \centering
    \includegraphics[width=0.6\textwidth]{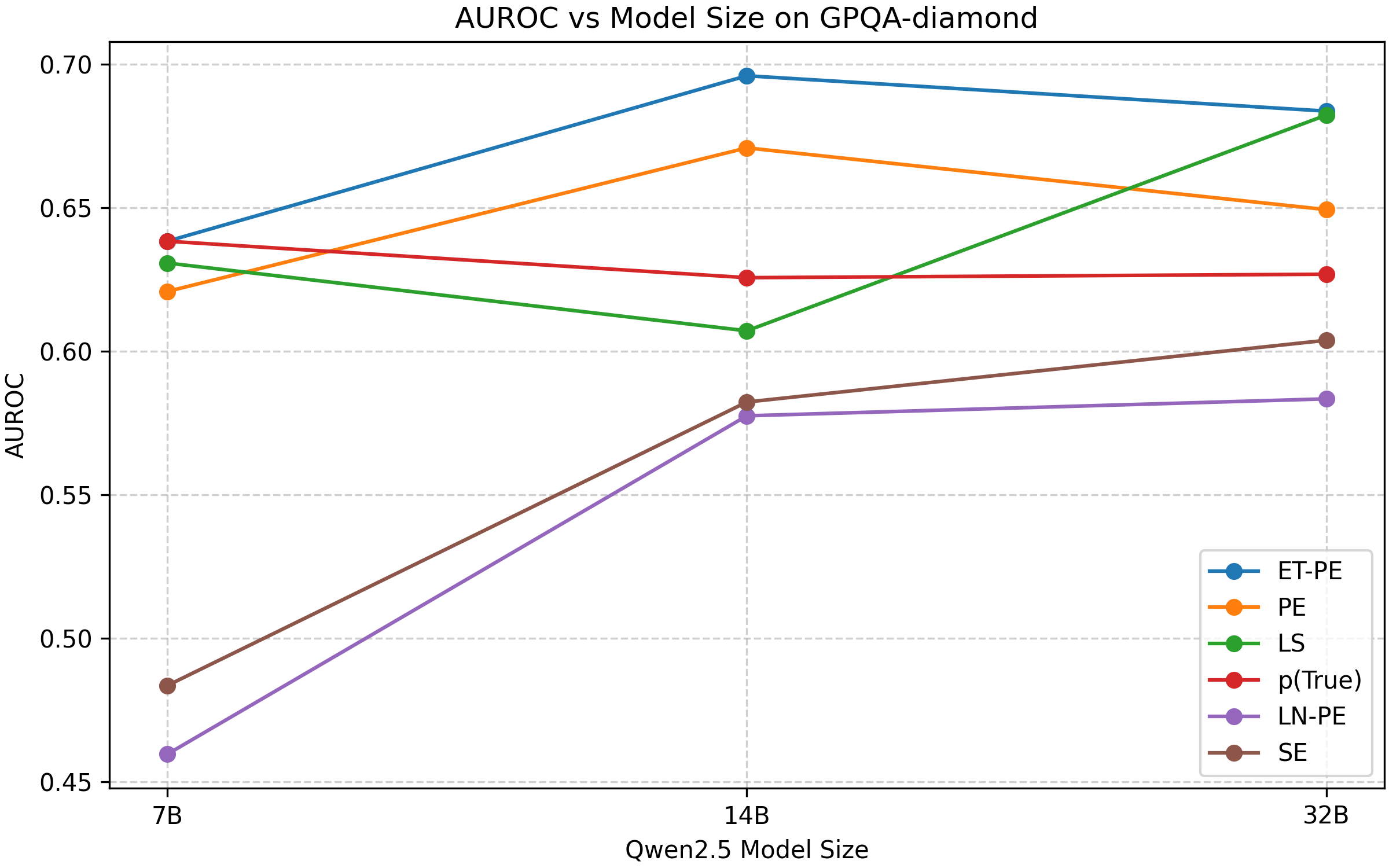}
    \caption{AUROC of different methods on the GPQA-diamond.}
    \label{fig:auroc}
\end{figure}

Figure~\ref{fig:auroc} shows the AUROC comparison for the Qwen2.5 series models on GPQA-diamond. Across models of different scales, the predictive entropy sampled by Entropy-Tree consistently demonstrates optimal performance. Additionally, semantic entropy becomes noticeably ineffective when dealing with long inference texts, whereas our method still holds an advantage over traditional predictive entropy. The results on all models and datasets are shown in Table~\ref{tab:auroc}, ET-PE (predictive entropy derived from Entropy-Tree results) demonstrates optimal calibration performance across various models and datasets. Although Entropy-Tree does not inherently propose a new uncertainty quantification metric, the model benefits from a more thorough exploration of ideas within the inference space due to the high-entropy-guided decoding strategy. As a result, the distribution of answers aligns better with the correctness of the model's responses.

\begin{table}[htbp]
\centering
\caption{AUROC comparison across models and datasets.}
\begin{tabularx}{\textwidth}{l c *{4}{X}}
\toprule
Model & Method & MATH-500 & SciBench & GPQA-main & GPQA-diamond \\
\midrule
\multirow{6}{*}{Qwen2.5-7B-Instruct}
& \textbf{ET-PE} & \textbf{0.840} & 0.785 & \textbf{0.590} & \textbf{0.638} \\
& PE & 0.806 & 0.787 & 0.586 & 0.621 \\
& LN-PE & 0.606 & 0.589 & 0.541 & 0.460 \\
& SE & 0.558 & 0.577 & 0.544 & 0.484 \\
& LS & 0.704 & 0.765 & 0.601 & 0.631 \\
& p(True) & 0.736 & 0.771 & 0.649 & 0.638 \\
\midrule
\multirow{6}{*}{Qwen2.5-14B-Instruct}
& \textbf{ET-PE} & \textbf{0.861} & \textbf{0.751} & 0.622 & \textbf{0.696} \\
& PE & 0.859 & 0.737 & 0.635 & 0.671 \\
& LN-PE & 0.693 & 0.622 & 0.526 & 0.578 \\
& SE & 0.661 & 0.630 & 0.532 & 0.582 \\
& LS & 0.852 & 0.753 & 0.613 & 0.607 \\
& p(True) & 0.657 & 0.620 & 0.586 & 0.626 \\
\midrule
\multirow{6}{*}{Qwen2.5-32B-Instruct}
& \textbf{ET-PE} & 0.751 & \textbf{0.767} & 0.636 & \textbf{0.684} \\
& PE & 0.751 & 0.765 & 0.672 & 0.649 \\
& LN-PE & 0.599 & 0.559 & 0.524 & 0.583 \\
& SE & 0.578 & 0.630 & 0.528 & 0.604 \\
& LS & 0.751 & 0.768 & 0.634 & 0.682 \\
& p(True) & 0.694 & 0.708 & 0.591 & 0.627 \\
\bottomrule
\end{tabularx}
\label{tab:auroc}
\end{table}

\subsection{Ablation Studies}

\begin{figure}[htbp]
    \centering
    \includegraphics[width=0.6\textwidth]{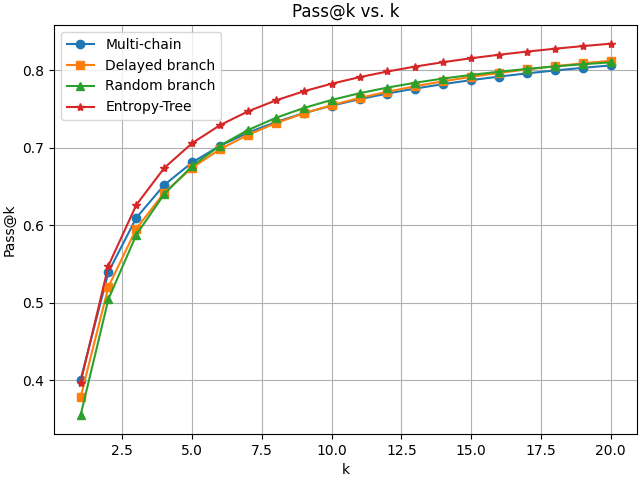}
    \caption{Results of ablation studies.}
    \label{fig:ablation}
\end{figure}

We conducted two ablation studies on the experimental performance of Qwen2.5-7B-Instruct on MATH-500, with the results shown in Figure~\ref{fig:ablation}.

\paragraph{Effect of Branching Position.}
Moving the entropy threshold from 80th to 90th percentile shifts branching points later in generation, resulting in performance degradation. This suggests that branching early in the decoding process may be more effective in improving decoding quality.

\paragraph{Effect of High-Entropy guided Branching.}
Random branching (without entropy guidance) at matched probability shows significantly worse performance than branching at high-entropy tokens, confirming the importance of high-entropy positions as ``decision forks''.

\section{Related Work}

\subsection{Decoding Strategies for Language Models}

\paragraph{Tree-based decoding strategies.}
Traditional tree-based decoding methods have been widely used in sequence generation tasks to explore and select optimal paths in a search tree. Beam search \cite{tillmann-ney-2003-word}, for instance, maintains a fixed number of promising hypotheses at each step, effectively pruning the search space while expanding nodes in a breadth-first manner. Monte Carlo Tree Search (MCTS) \cite{coulom2006efficient} extends this by incorporating probabilistic simulations and value estimations to guide exploration, balancing exploitation and exploration in complex decision spaces. Other approaches include variants like diverse beam search \cite{vijayakumar2018diversebeamsearchdecoding}, which introduces diversity penalties to encourage varied outputs, and A*-inspired heuristics \cite{lu-etal-2022-neurologic} that prioritize nodes based on estimated costs to the goal.

Recent advancements have introduced novel tree-based decoding strategies tailored for large language models (LLMs). TreeRL \cite{hou-etal-2025-treerl} leverages reinforcement learning to optimize tree exploration policies, enabling adaptive sampling and reward-based pruning. Similarly, TreePO \cite{li2025treepobridginggappolicy} employs policy optimization techniques within a tree structure to refine decoding paths, focusing on long-term sequence quality. 

\paragraph{Other decoding strategies.}
Traditional methods include greedy decoding \cite{jm3}, which selects the highest-probability token at each step for simplicity and speed, though it often leads to repetitive or suboptimal outputs. Sampling-based techniques, such as top-k sampling \cite{radford2019language} and nucleus (top-p) sampling \cite{holtzman2020curiouscaseneuraltext}, introduce randomness by restricting the selection to a subset of high-probability tokens, promoting diversity and creativity in generated text.

Recent innovations focus on accelerating inference. Speculative decoding \cite{leviathan2023fastinferencetransformersspeculative}, for example, employs a smaller draft model to predict multiple tokens ahead, which are then verified in parallel by the main LLM, significantly reducing latency in autoregressive generation. Lookahead decoding \cite{fu2024breaksequentialdependencyllm} builds on this by anticipating future tokens through heuristic approximations, enabling faster decoding while maintaining output quality. Other methods, like Medusa \cite{cai2024medusasimplellminference}, utilize multiple decoding heads to generate speculative branches, further optimizing for throughput in real-time applications.

\subsection{Uncertainty Quantification}

Classical approaches include Bayesian neural networks \cite{blundell2015weightuncertaintyneuralnetworks}, which place distributions over weights but suffer from computational intractability; MC dropout \cite{gal2016dropoutbayesianapproximationrepresenting}, which approximates Bayesian inference but requires many forward passes; and deep ensembles \cite{lakshminarayanan2017simplescalablepredictiveuncertainty}, which train multiple models at significant cost. Guo et al. \cite{guo2017calibrationmodernneuralnetworks} showed modern networks are poorly calibrated despite high accuracy. 

For LLMs, classical methods face fundamental limitations due to model scale. Kuhn et al. \cite{kuhn2023semanticuncertaintylinguisticinvariances} introduced semantic entropy, computing entropy over semantic equivalence classes. This was extended to hallucination detection \cite{quevedo2024detectinghallucinationslargelanguage}. Kadavath et al. \cite{kadavath2022languagemodelsmostlyknow} and Lin et al. \cite{lin2022teachingmodelsexpressuncertainty} explored querying models for confidence. However, Xiong et al. \cite{ICLR2024_6733cf15} demonstrated systematic overconfidence in verbalized uncertainty. For Black-box scenarios, SelfCheckGPT \cite{manakul-etal-2023-selfcheckgpt} detects hallucinations through sampling consistency. Lin et al. \cite{lin2024generatingconfidenceuncertaintyquantification} introduced semantic dispersion in embedding space. Conformal prediction methods \cite{pmlr-v235-mohri24a,NEURIPS2024_d02ff1ae} provide statistical guarantees but require calibration data. 

\section{Conclusion}

In this paper, we point out that traditional decoding strategies are blind in diversity exploration. We propose Entropy-Tree decoding, which focuses on tokens with high entropy and certain semantic importance during the decoding process, branching out from these tokens to allow the model to actively explore new methods and strategies. Our experiments demonstrate that Entropy-Tree exhibits outstanding performance in improving the accuracy and calibration of multiple sampling.

\section{Limitations and Future Work}

We acknowledge that the current approach has some limitations. Our method requires access to token logits and self-attention scores, which can introduce additional latency. Moreover, some APIs do not support access to model logits or self-attention scores, potentially limiting the application of our method in black-box scenarios. 

In the future, we plan to further expand upon the existing approach by integrating it into current LLM inference engines (such as vLLM \cite{kwon2023efficientmemorymanagementlarge}) and investigating the impact of our method on improving the efficiency and convergence speed of reinforcement learning sampling.


\end{document}